\documentclass[11pt]{article}
\usepackage{acl2014}
\usepackage{times}
\usepackage{latexsym}
\usepackage{pstricks}
\usepackage{rtrees}
\usepackage{graphicx}
\usepackage{amsmath}

\title{Strongly Incremental Repair Detection}

\author{Julian Hough$^{1,2}$ \\
 $^{1}$Dialogue Systems Group \\
  Faculty of Linguistics\\and Literature\\
  Bielefeld University\\
  {\tt julian.hough@uni-bielefeld.de} \\\And
  Matthew Purver$^{2}$  \\
  $^{2}$Cognitive Science Research Group\\
  School of Electronic Engineering\\and Computer Science\\
  Queen Mary University of London \\
  {\tt m.purver@qmul.ac.uk} \\}

\date{}

\begin{document}
\maketitle
\begin{abstract}
We present STIR (STrongly Incremental Repair detection), a system that detects speech repairs and edit terms on transcripts incrementally with minimal latency. STIR uses information-theoretic measures from n-gram models as its principal decision features in a pipeline of classifiers detecting the different stages of repairs. Results on the Switchboard disfluency tagged corpus show utterance-final accuracy on a par with state-of-the-art incremental repair detection methods, but with better incremental accuracy, faster time-to-detection and less computational overhead. We evaluate its performance using incremental metrics and propose new repair processing evaluation standards.
\end{abstract}

\section{Introduction}

Self-repairs in spontaneous speech are annotated according to a well established three-phase structure from \cite{Shriberg94} onwards, and as described in \newcite{Meeter.etal95}'s Switchboard corpus annotation handbook:

\vspace{-0.3cm}
\begin{equation}\label{eg:SRannotationStructure}
\strut \text{ John }
\underbrace{\strut \text{ [ likes } }_\text{reparandum}+
\underbrace{\strut \text{ \{uh\}} }_\text{interregnum}
\underbrace{\strut \text{  loves ]} }_\text{repair}
\strut \text{ Mary}
\end{equation}
\vspace{-0.3cm}

\noindent
From a dialogue systems perspective, detecting repairs and assigning them the appropriate structure is vital for robust natural language understanding (NLU) in interactive systems. Downgrading the commitment of \emph{reparandum} phases and assigning appropriate \emph{interregnum} and \emph{repair} phases permits computation of the user's intended meaning. 

Furthermore, the recent focus on \emph{incremental} dialogue systems (see e.g. \cite{Rieser.Schlangen11}) means that repair detection should operate without unnecessary processing overhead, and function efficiently within an incremental framework. 
However, such left-to-right operability on its own is not sufficient:
in line with the principle of strong incremental interpretation \cite{Milward91}, a repair detector should give \emph{the best results possible as early as possible}. With one exception \cite{Zwarts.etal10}, there has been no focus on evaluating or improving the \emph{incremental performance} of repair detection.


In this paper we present STIR (Strongly Incremental Repair detection), a system which addresses the challenges of incremental accuracy, computational complexity and latency in self-repair detection, by making local decisions based on relatively simple measures of fluency and similarity. Section \ref{sec:previous} reviews state-of-the-art methods; Section~\ref{sec:approach} summarizes the challenges and explains our general approach; Section~\ref{sec:STIR} explains STIR in detail; Section~\ref{sec:experiment} explains our experimental set-up and novel evaluation metrics; Section~\ref{sec:results} presents and discusses our results and Section~\ref{sec:conclusion} concludes.

\section{Previous work}
\label{sec:previous}

%

\newcite{Qian.Liu13} achieve the state of the art in Switchboard
corpus self-repair detection, with an F-score for detecting reparandum
words of 0.841 using a three-step weighted Max-Margin Markov network
approach. Similarly, \newcite{Georgila09} uses Integer Linear
Programming post-processing of a CRF to achieve F-scores over
0.8 for reparandum start and repair start detection. However neither approach can operate incrementally.

Recently, there has been increased interest in left-to-right repair detection: \newcite{Rasooli.Tetreault14} and \newcite{Honnibal.Johnson14} present dependency parsing systems with reparandum detection which perform similarly, the latter equalling \newcite{Qian.Liu13}'s F-score at 0.841. However, while operating left-to-right, these systems are not designed or evaluated for their \emph{incremental} performance. The use of beam search over different repair hypotheses in \cite{Honnibal.Johnson14} is likely to lead to unstable repair label sequences, and they report repair hypothesis `jitter'. Both of these systems use a non-monotonic dependency parsing approach that immediately removes the reparandum from the linguistic analysis of the utterance in terms of its dependency structure and repair-reparandum correspondence, which from a downstream NLU module's perspective is undesirable. \newcite{Heeman.Allen99} and \newcite{Miller.Schuler08} present earlier left-to-right operational detectors which are less accurate and again give no indication of the incremental performance of their systems. While \newcite{Heeman.Allen99} rely on repair structure template detection coupled with a multi-knowledge-source language model, the rarity of the tail of repair structures is likely to be the reason for lower performance: \newcite{Hough.Purver13SemDial} show that only 39\% of repair alignment structures appear at least twice in Switchboard, supported by the 29\% reported by \newcite{Heeman.Allen99} on the smaller TRAINS corpus. \newcite{Miller.Schuler08}'s encoding of repairs into a grammar also causes sparsity in training: repair is a general processing strategy not restricted to certain lexical items or POS tag sequences.

The model we consider most suitable for incremental dialogue systems so far is \newcite{Zwarts.etal10}'s incremental version of \newcite{Johnson.Charniak04}'s noisy channel repair detector, as it incrementally applies structural repair analyses (rather than just identifying reparanda) and is evaluated for its incremental properties. Following \cite{Johnson.Charniak04}, their system uses an n-gram language model trained on roughly 100K utterances of reparandum-excised (`cleaned') Switchboard data. Its channel model is a statistically-trained S-TAG parser whose grammar has simple reparandum-repair alignment rule categories for its non-terminals (copy, delete, insert, substitute) and words for its terminals. The parser hypothesises all possible repair structures for the string consumed so far in a chart, before pruning the unlikely ones. It performs equally well to the non-incremental model by the end of each utterance (F-score = 0.778), and can make detections early via the addition of a speculative next-word repair completion category to their S-TAG non-terminals. In terms of incremental performance, they report the novel evaluation metric of \emph{time-to-detection} for correctly identified repairs, achieving an average of 7.5 words from the start of the reparandum and 4.6 from the start of the repair phase. They also introduce \emph{delayed accuracy}, a word-by-word evaluation against gold-standard disfluency tags up to the word before the current word being consumed (in their terms, the \emph{prefix boundary}), giving a measure of the stability of the repair hypotheses. They report an F-score of 0.578 at one word back from the current prefix boundary, increasing word-by-word until 6 words back where it reaches 0.770. These results are the point-of-departure for our work.

\section{Challenges and Approach}
\label{sec:approach}

In this section we summarize the challenges for incremental repair detection: computational complexity, repair hypothesis stability, latency of detection and repair structure identification. In \ref{sec:newapproach} we explain how we address these.

\paragraph{Computational complexity} Approaches to detecting repair structures often use chart storage \cite{Zwarts.etal10,Johnson.Charniak04,Heeman.Allen99}, which poses a computational overhead: if considering all possible boundary points for a repair structure's 3 phases beginning on any word, for prefixes of length $n$ the number of hypotheses can grow in the order $O(n^{4})$. Exploring a subset of this space is necessary for assigning entire repair structures as in (\ref{eg:SRannotationStructure}) above, rather than just detecting reparanda: the \cite{Johnson.Charniak04,Zwarts.etal10} noisy-channel detector is the only system that applies such structures but the potential run-time complexity in decoding these with their S-TAG repair parser is $O(n^{5})$. In their approach, complexity is mitigated by imposing a maximum repair length (12 words), and also by using beam search with re-ranking \cite{Lease.etal06,Zwarts.Johnson11LMs}. If we wish to include full decoding of the repair's structure (as argued by \newcite{Hough.Purver13SemDial} as necessary for full interpretation) whilst taking a strictly incremental and time-critical perspective, reducing this complexity by minimizing the size of this search space is crucial.



\paragraph{Stability of repair hypotheses and latency}
Using a beam search of n-best hypotheses on a word-by-word basis can cause `jitter' in the detector's output. While utterance-final accuracy is desired, for a truly incremental system good intermediate results are equally important. \newcite{Zwarts.etal10}'s time-to-detection results show their system is only certain about a detection after processing the entire repair. This may be due to the string alignment-inspired S-TAG that matches repair and reparanda: a `rough copy' dependency only becomes likely once the entire repair has been consumed. The latency of 4.6 words to detection and a relatively slow rise to utterance-final accuracy up to 6 words back is undesirable given repairs have a mean reparandum length of $\approx$1.5 words \cite{Hough.Purver13SemDial,Shriberg.Stolcke98}.


\paragraph{Structural identification} Classifying repairs has been ignored in repair processing, despite the presence of distinct categories (e.g. repeats, substitutions, deletes) with different pragmatic effects  \cite{Hough.Purver13SemDial}.\footnote{Though see \cite{Germesin.etal08} for one approach, albeit using idiosyncratic repair categories.} This is perhaps due to lack of clarity in definition: even for human annotators, verbatim repeats withstanding, agreement is often poor \cite{Hough.Purver13SemDial,Shriberg94}. Assigning and evaluating repair (not just reparandum) structures will allow repair interpretation in future; however, work to date evaluates only reparandum detection.


\newcommand{\given}[0]{\!\!\mid\!\!}

\subsection{Our approach}
\label{sec:newapproach}
To address the above, we propose an alternative to \cite{Johnson.Charniak04,Zwarts.etal10}'s noisy channel model. While the model elegantly captures intuitions about parallelism in repairs and modelling fluency, it relies on string-matching, motivated in a similar way to automatic spelling correction \cite{Brill.Moore00}: it assumes a speaker chooses to utter fluent utterance $X$ according to some prior distribution $P(X)$, but a noisy channel causes them instead to utter a noisy $Y$  according to channel model $P(Y\given X)$. Estimating $P(Y\given X)$ directly from observed data is difficult due to sparsity of repair instances, so a transducer is trained on the rough copy alignments between reparandum and repair. This approach succeeds because repetition and simple substitution repairs are very common; but repair as a psychological process is not driven by string alignment, and deletes, restarts and rarer substitution forms are not captured. Furthermore, the noisy channel model assumes an inherently utterance-global process for generating (and therefore finding) an underlying `clean' string --- much as similar spelling correction models are word-global --- we instead take a very local perspective here.


In accordance with psycholinguistic evidence \cite{Brennan.Schober01}, we assume characteristics of the repair onset allow hearers to detect it very quickly and solve the \emph{continuation problem} \cite{Levelt83} of integrating the repair into their linguistic context immediately, before processing or even hearing the end of the repair phase. While repair onsets may take the form of interregna, this is not a reliable signal, occurring in only $\approx$15\% of repairs \cite{Hough.Purver13SemDial,Heeman.Allen99}. Our repair onset detection is therefore driven by departures from fluency, via information-theoretic features derived incrementally from a language model in line with recent psycholinguistic accounts of incremental parsing -- see \cite{Keller04,Jaeger.Tily11}.
 
Considering the time-linear way a repair is processed and the fact speakers are exponentially less likely to trace one word further back in repair as utterance length increases \cite{Shriberg.Stolcke98}, backwards search seems to be the most efficient reparandum extent detection method.\footnote{We acknowledge a purely position-based model for reparandum extent detection under-estimates prepositions, which speakers favour as the retrace start and over-estimates verbs, which speakers tend to avoid retracing back to, preferring to begin the utterance again, as \cite{Healey.etal11}'s experiments also demonstrate.} Features determining the detection of the reparandum extent in the backwards search can also be information-theoretic: entropy measures of distributional parallelism can characterize not only rough copy dependencies, but distributionally similar or dissimilar correspondences between sequences. Finally, when detecting the repair end and structure, distributional information allows computation of the similarity between reparandum and repair. We argue a local-detection-with-backtracking approach is more cognitively plausible than string-based left-to-right repair labelling, and using this insight should allow an improvement in incremental accuracy, stability and time-to-detection over string-alignment driven approaches in repair detection.

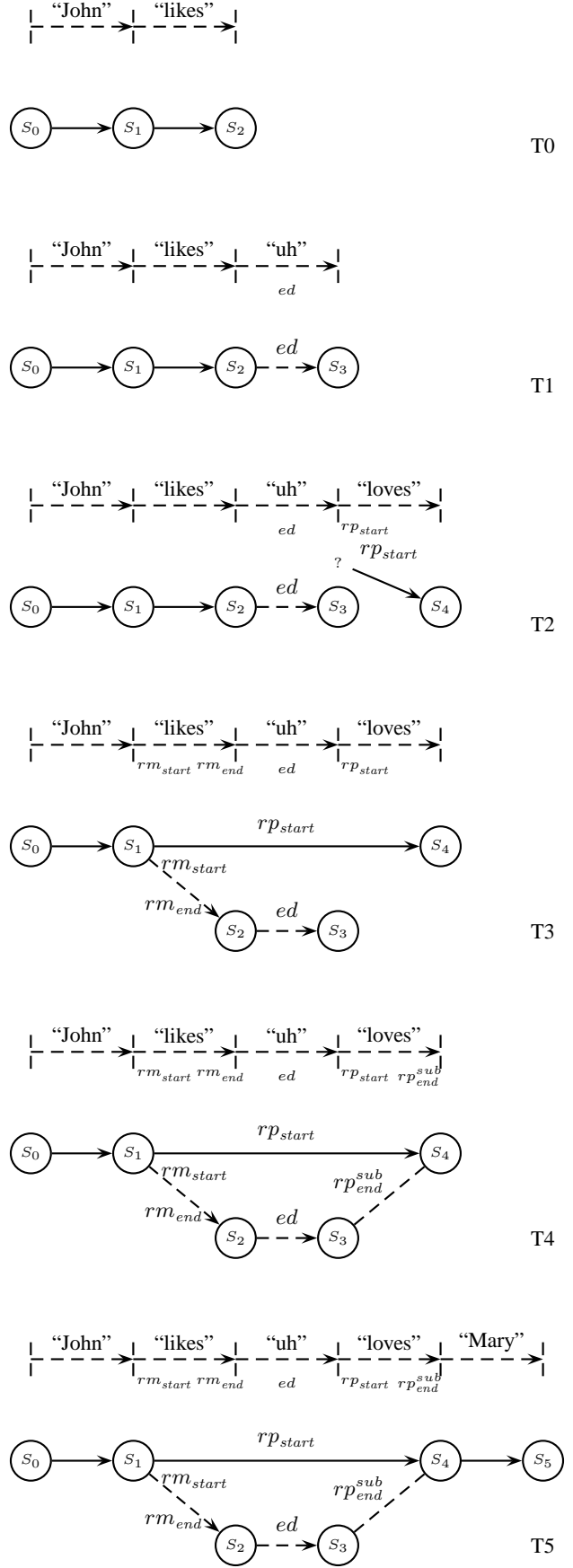
\begin{figure}
\vspace*{-0.3cm}
\begin{small}

\begin{pspicture}(8,3.5)
\psset{arrows=->,arrowscale=1.5}
\psline[linestyle=dashed,arrows=-](0,2.5)(0,3)
\psline[linestyle=dashed,arrows=->](0,2.75)(1.5,2.75)
\psline[linestyle=dashed,arrows=-](1.5,2.5)(1.5,3)
\rput(0.75,3){``John''}
\psline[linestyle=dashed,arrows=->](1.5,2.75)(3,2.75)
\psline[linestyle=dashed,arrows=-](3,2.5)(3,3)
\rput(2.25,3){``likes''}

\begin{tiny}
\cnodeput(0,1.25){S0}{$S_0$}
\cnodeput(1.5,1.25){S1}{$S_1$}\ncline{->}{S0}{S1}
\cnodeput(3,1.25){S2}{$S_2$}\ncline{->}{S1}{S2}\end{tiny} 
\rput(7.5,1){T0}
\end{pspicture}

\begin{pspicture}(8,3.5)
\psset{arrows=->,arrowscale=1.5}
\psline[linestyle=dashed,arrows=-](0,2.5)(0,3)
\psline[linestyle=dashed,arrows=->](0,2.75)(1.5,2.75)
\psline[linestyle=dashed,arrows=-](1.5,2.5)(1.5,3)
\rput(0.75,3){``John''}
\psline[linestyle=dashed,arrows=->](1.5,2.75)(3,2.75)
\psline[linestyle=dashed,arrows=-](3,2.5)(3,3)
\rput(2.25,3){``likes''}
\psline[linestyle=dashed,arrows=->](3,2.75)(4.5,2.75)
\psline[linestyle=dashed,arrows=-](4.5,2.5)(4.5,3)
\rput(3.75,3){``uh''}
\begin{tiny}\rput(3.75,2.4){$ed$}\end{tiny}

\begin{tiny}
\cnodeput(0,1.25){S0}{$S_0$}
\cnodeput(1.5,1.25){S1}{$S_1$}\ncline{->}{S0}{S1}
\cnodeput(3,1.25){S2}{$S_2$}\ncline{->}{S1}{S2}
\cnodeput(4.5,1.25){S3}{$S_3$}\ncline[linestyle=dashed,arrows=->]{->}{S2}{S3}\end{tiny}\taput{$ed$}
\rput(7.5,1){T1}
\end{pspicture}

\begin{pspicture}(7.5,3.5)
\psset{arrows=->,arrowscale=1.5}
\psline[linestyle=dashed,arrows=-](0,2.5)(0,3)
\psline[linestyle=dashed,arrows=->](0,2.75)(1.5,2.75)
\psline[linestyle=dashed,arrows=-](1.5,2.5)(1.5,3)
\rput(0.75,3){``John''}
\psline[linestyle=dashed,arrows=->](1.5,2.75)(3,2.75)
\psline[linestyle=dashed,arrows=-](3,2.5)(3,3)
\rput(2.25,3){``likes''}
\psline[linestyle=dashed,arrows=->](3,2.75)(4.5,2.75)
\psline[linestyle=dashed,arrows=-](4.5,2.5)(4.5,3)
\rput(3.75,3){``uh''}
\begin{tiny}\rput(3.75,2.4){$ed$}\end{tiny}
\psline[linestyle=dashed,arrows=->](4.5,2.75)(6,2.75)
\psline[linestyle=dashed,arrows=-](6,2.5)(6,3)
\rput(5.25,3){``loves''}
\begin{tiny}\rput(4.9,2.4){$rp_{\mathit{start}}^{}$}\end{tiny}

\begin{tiny}
\cnodeput(0,1.25){S0}{$S_0$}
\cnodeput(1.5,1.25){S1}{$S_1$}\ncline{->}{S0}{S1}
\cnodeput(3,1.25){S2}{$S_2$}\ncline{->}{S1}{S2}
\cnodeput(4.5,1.25){S3}{$S_3$}\ncline[linestyle=dashed,arrows=->]{->}{S2}{S3}\end{tiny}\taput{$ed$}\begin{tiny}
\cnodeput[linecolor=white](4.5,1.9){S121}{$?$}
\cnodeput(6,1.25){S4}{$S_4$}\ncline{->}{S121}{S4}\end{tiny}\taput{$rp_{\mathit{start}}^{}$}
\rput(7.5,1){T2}
\end{pspicture}

\begin{pspicture}(7.5,3.5)
\psset{arrows=->,arrowscale=1.5}
\psline[linestyle=dashed,arrows=-](0,2.5)(0,3)
\psline[linestyle=dashed,arrows=->](0,2.75)(1.5,2.75)
\psline[linestyle=dashed,arrows=-](1.5,2.5)(1.5,3)
\rput(0.75,3){``John''}
\psline[linestyle=dashed,arrows=->](1.5,2.75)(3,2.75)
\psline[linestyle=dashed,arrows=-](3,2.5)(3,3)
\rput(2.25,3){``likes''}
\begin{tiny}\rput(1.96,2.4){$rm_{\mathit{start}}^{}$}\rput(2.78,2.4){$rm_{\mathit{end}}^{}$}\end{tiny}
\psline[linestyle=dashed,arrows=->](3,2.75)(4.5,2.75)
\psline[linestyle=dashed,arrows=-](4.5,2.5)(4.5,3)
\rput(3.75,3){``uh''}
\begin{tiny}\rput(3.75,2.4){$ed$}\end{tiny}
\psline[linestyle=dashed,arrows=->](4.5,2.75)(6,2.75)
\psline[linestyle=dashed,arrows=-](6,2.5)(6,3)
\rput(5.25,3){``loves''}
\begin{tiny}\rput(4.9,2.4){$rp_{\mathit{start}}^{}$}\end{tiny}

\begin{tiny}
\cnodeput(0,1.25){S0}{$S_0$}
\cnodeput(1.5,1.25){S1}{$S_1$}\ncline{->}{S0}{S1}
\cnodeput(3,0){S2}{$S_2$}\ncline[linestyle=dashed,arrows=-]{->}{S1}{S2}\end{tiny}\rput(2.4,0.95){$rm_{\mathit{start}}^{}$}\rput(2.1,0.35){$rm_{\mathit{end}}^{}$}\begin{tiny}
\cnodeput(4.5,0){S3}{$S_3$}\ncline[linestyle=dashed,arrows=->]{->}{S2}{S3}\end{tiny}\taput{$ed$}\begin{tiny}
\cnodeput(6,1.25){S4}{$S_4$}\ncline{->}{S1}{S4}\end{tiny}\taput{$rp_{\mathit{start}}^{}$}
\rput(7.5,0){T3}
\end{pspicture}

\vspace{1cm}
\begin{pspicture}(7.5,3.5)
\psset{arrows=->,arrowscale=1.5}
\psline[linestyle=dashed,arrows=-](0,2.5)(0,3)
\psline[linestyle=dashed,arrows=->](0,2.75)(1.5,2.75)
\psline[linestyle=dashed,arrows=-](1.5,2.5)(1.5,3)
\rput(0.75,3){``John''}
\psline[linestyle=dashed,arrows=->](1.5,2.75)(3,2.75)
\psline[linestyle=dashed,arrows=-](3,2.5)(3,3)
\rput(2.25,3){``likes''}
\begin{tiny}\rput(1.96,2.4){$rm_{\mathit{start}}^{}$}\rput(2.78,2.4){$rm_{\mathit{end}}^{}$}\end{tiny}
\psline[linestyle=dashed,arrows=->](3,2.75)(4.5,2.75)
\psline[linestyle=dashed,arrows=-](4.5,2.5)(4.5,3)
\rput(3.75,3){``uh''}
\begin{tiny}\rput(3.75,2.4){$ed$}\end{tiny}
\psline[linestyle=dashed,arrows=->](4.5,2.75)(6,2.75)
\psline[linestyle=dashed,arrows=-](6,2.5)(6,3)
\rput(5.25,3){``loves''}
\begin{tiny}\rput(4.9,2.4){$rp_{\mathit{start}}^{}$}\rput(5.7,2.4){$rp_{\mathit{end}}^{sub}$}\end{tiny}

\begin{tiny}
\cnodeput(0,1.25){S0}{$S_0$}
\cnodeput(1.5,1.25){S1}{$S_1$}\ncline{->}{S0}{S1}
\cnodeput(3,0){S2}{$S_2$}\ncline[linestyle=dashed,arrows=-]{->}{S1}{S2}\end{tiny}\rput(2.4,0.95){$rm_{\mathit{start}}^{}$}\rput(2.1,0.35){$rm_{\mathit{end}}^{}$}\begin{tiny}
\cnodeput(4.5,0){S3}{$S_3$}\ncline[linestyle=dashed,arrows=->]{->}{S2}{S3}\end{tiny}\taput{$ed$}\begin{tiny}
\cnodeput(6,1.25){S4}{$S_4$}\ncline{->}{S1}{S4}\end{tiny}\taput{$rp_{\mathit{start}}^{}$}
\begin{tiny}
\ncline[linestyle=dashed,arrows=-]{-}{S3}{S4}\end{tiny}\rput(4.8,0.85){$rp_{\mathit{end}}^{sub}$}
\rput(7.5,0){T4}
\end{pspicture}


\vspace{1cm}
\begin{pspicture}(7.5,3.5)
\psset{arrows=->,arrowscale=1.5}
\psline[linestyle=dashed,arrows=-](0,2.5)(0,3)
\psline[linestyle=dashed,arrows=->](0,2.75)(1.5,2.75)
\psline[linestyle=dashed,arrows=-](1.5,2.5)(1.5,3)
\rput(0.75,3){``John''}
\psline[linestyle=dashed,arrows=->](1.5,2.75)(3,2.75)
\psline[linestyle=dashed,arrows=-](3,2.5)(3,3)
\rput(2.25,3){``likes''}
\begin{tiny}\rput(1.96,2.4){$rm_{\mathit{start}}^{}$}\rput(2.78,2.4){$rm_{\mathit{end}}^{}$}\end{tiny}
\psline[linestyle=dashed,arrows=->](3,2.75)(4.5,2.75)
\psline[linestyle=dashed,arrows=-](4.5,2.5)(4.5,3)
\rput(3.75,3){``uh''}
\begin{tiny}\rput(3.75,2.4){$ed$}\end{tiny}
\psline[linestyle=dashed,arrows=->](4.5,2.75)(6,2.75)
\psline[linestyle=dashed,arrows=-](6,2.5)(6,3)
\rput(5.25,3){``loves''}
\begin{tiny}\rput(4.9,2.4){$rp_{\mathit{start}}^{}$}\rput(5.7,2.4){$rp_{\mathit{end}}^{sub}$}\end{tiny}
\psline[linestyle=dashed,arrows=->](6,2.75)(7.5,2.75)
\psline[linestyle=dashed,arrows=-](7.5,2.5)(7.5,3)
\rput(6.75,3){``Mary''}

\begin{tiny}
\cnodeput(0,1.25){S0}{$S_0$}
\cnodeput(1.5,1.25){S1}{$S_1$}\ncline{->}{S0}{S1}
\cnodeput(3,0){S2}{$S_2$}\ncline[linestyle=dashed,arrows=-]{->}{S1}{S2}\end{tiny}\rput(2.4,0.95){$rm_{\mathit{start}}^{}$}\rput(2.1,0.35){$rm_{\mathit{end}}^{}$}\begin{tiny}
\cnodeput(4.5,0){S3}{$S_3$}\ncline[linestyle=dashed,arrows=->]{->}{S2}{S3}\end{tiny}\taput{$ed$}\begin{tiny}
\cnodeput(6,1.25){S4}{$S_4$}\ncline{->}{S1}{S4}\end{tiny}\taput{$rp_{\mathit{start}}^{}$}
\ncline[linestyle=dashed,arrows=-]{-}{S3}{S4}\rput(4.8,0.85){$rp_{\mathit{end}}^{sub}$}\begin{tiny}
\cnodeput(7.5,1.25){S5}{$S_5$}\end{tiny}\ncline{->}{S4}{S5}\taput{}
\rput(7.5,0){T5}
\end{pspicture}

\end{small}

\caption{Strongly Incremental Repair Detection}\label{fig:STIR}
\end{figure}

\section{STIR: Strongly Incremental Repair detection}
\label{sec:STIR}
Our system, STIR (Strongly Incremental Repair detection), therefore takes a local incremental approach to detecting repairs and isolated edit terms, assigning words the structures in (\ref{eq:detectionStructure}). We include interregnum recognition in the process, due to the inclusion of interregnum vocabulary within edit term vocabulary \cite{Ginzburg12,Hough.Purver13SemDial}, a useful feature for repair detection \cite{Lease.etal06,Qian.Liu13}.

\begin{small}
\begin{align}
\begin{cases}{}
... [ rm_{\mathit{start}} ... rm_{\mathit{end}} +  \{ed\} rp_{\mathit{start}} ... rp_{\mathit{end}} ] ... & \\ 
... \{ed\}... &
 \end{cases}\label{eq:detectionStructure}
\end{align}
\end{small}

\noindent
Rather than detecting the repair structure in its left-to-right string order as above, STIR functions as in Figure~\ref{fig:STIR}: first detecting edit terms (possibly interregna) at step T1; then detecting repair onsets $rp_{\mathit{start}}$ at T2; if one is found, backwards searching to find  $rm_{\mathit{start}}$ at T3; then finally finding the repair end $rp_{\mathit{end}}$ at T4. Step T1 relies mainly on lexical probabilities from an edit term language model; T2 exploits features of divergence from a fluent language model; T3 uses fluency of hypothesised repairs; and T4 the similarity between distributions after reparandum and repair. However, each stage integrates these basic insights via multiple related features in a statistical classifier.

\subsection{Enriched incremental language models}
\label{sec:LMs}

We derive the basic information-theoretic features required using
n-gram language models, as they have a long history of information
theoretic analysis \cite{Shannon48} and provide reproducible results
without forcing commitment to one particular grammar
formalism. Following recent work on modelling grammaticality
judgements \cite{Clark.etal13CMCL}, we implement several modifications
to standard language models to develop our basic measures of fluency
and uncertainty.



For our main fluent language models we train a trigram model with Kneser-Ney smoothing \cite{Kneser.Ney95} on the words and POS tags of the standard Switchboard training data (all files with conversation numbers beginning sw2*,sw3* in the Penn Treebank III release), consisting of $\approx$100K utterances, $\approx$600K words. We follow \cite{Johnson.Charniak04} by cleaning the data of disfluencies (i.e. edit terms and reparanda), to approximate a `fluent' language model. We call these probabilities $p^{\mathit{lex}}_{kn}$,  $p^{\mathit{pos}}_{kn}$ below.\footnote{We suppress the $^{\mathit{pos}}$ and $^{\mathit{lex}}$ superscripts below where we refer to measures from either model.}

We then derive \emph{surprisal} as our principal default lexical uncertainty measurement $s$ (equation~\ref{eqPsyn}) in both models; and, following \cite{Clark.etal13CMCL}, the (unigram) Weighted Mean Log trigram probability (WML, eq.~\ref{eqWML})-- the trigram logprob of the sequence divided by the inverse summed logprob of the component unigrams (apart from the first two words in the sequence, which serve as the first trigram history). As here we use a local approach we restrict the WML measures to single trigrams (weighted by the inverse logprob of the final word). While use of standard n-gram probability conflates syntactic with lexical probability, WML gives us an approximation to \emph{incremental syntactic probability} by factoring out lexical frequency.

\vspace{-0.3cm}
\begin{small}
\begin{equation}
s(w_{i-2}\ldots w_i) = - \log_2 p_{kn}(w_i\mid w_{i-2},w_{i-1})~~
\label{eqPsyn}
\end{equation}
\begin{equation}
\hspace*{-0.25cm}\mathit{WML}(w_{0}\ldots w_{n}) = \frac{\sum_{i=2}^{i=n}\log_2 p_{kn}(w_i\mid w_{i-2}, w_{i-1})}{- \sum_{j=2}^{n} \log_2 p_{kn}(w_{j})}~~ 
\label{eqWML}
\end{equation}
\end{small}
\vspace{-0.3cm}


\paragraph{Distributional measures} To approximate uncertainty, we also derive the entropy $H(w\given c)$ of the possible word continuations $w$ given a context $c$, from $p(w_i\given c)$ for all words $w_i$ in the vocabulary -- see (\ref{eq:entropy1}). Calculating distributions over the entire lexicon incrementally is costly, so we approximate this by constraining the calculation to words which are observed at least once in context $c$ in training, $w_c = \{w | count(c, w) \geq 1\}$~,  assuming a uniform distribution over the unseen suffixes by using the appropriate smoothing constant, and subtracting the latter from the former -- see eq. (\ref{eq:entropy2}).

Manual inspection showed this approximation to be very close, and the trie structure of our n-gram models allows efficient calculation.  We also make use of the Zipfian distribution of n-grams in corpora by storing entropy values for the 20\% most common trigram contexts observed in training, leaving entropy values of rare or unseen contexts to be computed at decoding time with little search cost due to their small or empty $w_c$ sets.

\vspace{-0.3cm}
\begin{small}
\begin{equation}
H(w \mid c)  = -\sum_{w \in Vocab}p_{kn}(w \mid c) \log_2 p_{kn}(w\mid c)
\label{eq:entropy1}
\end{equation}

\begin{equation}
\begin{split}
H(w \mid c)  \approx &\left[-\sum_{w \in w_c} p_{kn}(w \mid c) \log_2 p_{kn}(w\mid c)\right] \\
& - \left[n \times \lambda \log_2\lambda \right]\\ \vspace*{0.3cm}
&  \text{where } n  = \left\vert{Vocab}\right\vert - \left\vert w_c\right\vert  \\
&  \text{and } \lambda = \frac{1 - \sum_{w \in w_c} p_{kn}(w \mid c)}{n} 
\end{split}\label{eq:entropy2}
\end{equation}
\end{small}
\vspace{-0.4cm}

Given entropy estimates, we can also similarly approximate the Kullback-Leibler (KL) divergence (relative entropy) between distributions in two different contexts $c_1$ and $c_2$, i.e. $\theta(w\vert c_1)$ and  $\theta(w\vert c_2)$, by pair-wise computing $p(w\vert c_1) \log_2( \frac{p(w \vert c_1)}{p(w \vert c_2)})$ only for words \begin{small} $w \in w_{c_1} \cap w_{c_2}$\end{small}~, then approximating  unseen values by assuming uniform distributions. Using $p_{kn}$ smoothed estimates rather than raw maximum likelihood estimations avoids infinite KL divergence values. Again, we found this approximation sufficiently close to the real values for our purposes. 
All such probability and distribution values are stored in incrementally constructed directed acyclic graph (DAG) structures (see Figure~\ref{fig:STIR}), exploiting the Markov assumption of n-gram models to allow efficient calculation by avoiding re-computation.

\begin{figure*}
\includegraphics[width=1.9\columnwidth]{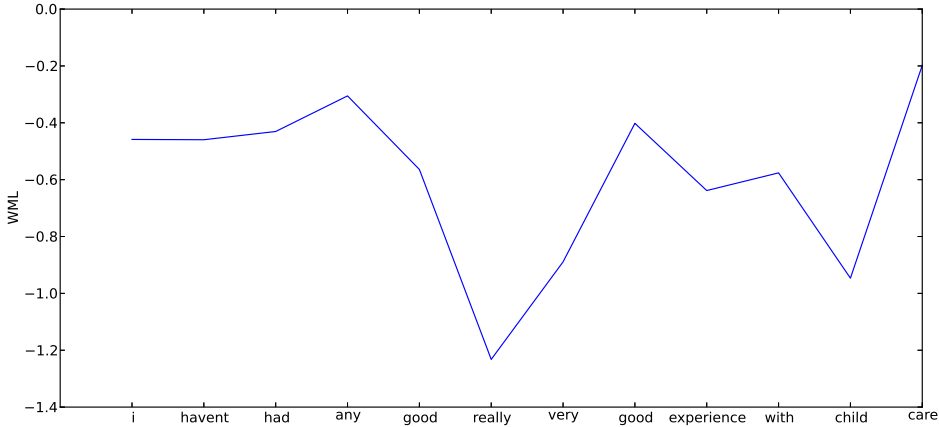}
\caption{WML measures for trigrams for a repaired utterance exhibiting the drop at the repair onset}\label{fig:WMLplot}
\end{figure*}

\subsection{Individual classifiers}

This section details the features used by the 4 individual classifiers. To investigate the utility of the features used in each classifier we obtain values on the standard Switchboard heldout data (PTB III files sw4[5-9]*: 6.4K utterances, 49K words).

\subsubsection{Edit term detection}
In the first component, we utilise the well-known observation that edit terms have a distinctive vocabulary \cite{Ginzburg12}, training a bigram model on a corpus of all edit words annotated in Switchboard's training data. The classifier simply uses the surprisal $s^{\mathit{lex}}$ from this edit word model, and the trigram surprisal $s^{\mathit{lex}}$ from the standard fluent model of Section~\ref{sec:LMs}. At the current position $w_n$, one, both or none of words $w_{n}$ and $w_{n-1}$ are classified as edits. We found this simple approach effective and stable, although some delayed decisions occur in cases where $s^{\mathit{lex}}$ and $\mathit{WML}^{\mathit{lex}}$ are high in both models before the end of the edit, e.g. ``I like" $\rightarrow$ ``I $\{like\}$ want...''. Words classified as $ed$ are removed from the incremental processing graph (indicated by the dotted line transition in Figure \ref{fig:STIR}) and the stack updated if repair hypotheses are cancelled due to a delayed edit hypothesis of $w_{n-1}$.

\subsubsection{Repair start detection}

Repair onset detection is arguably the most crucial component: the greater its accuracy, the better the input for downstream components and the lesser the overhead of filtering false positives required. We use Section~\ref{sec:LMs}'s information-theoretic features $s, \mathit{WML}, H$ for words and POS, and introduce 5 additional information-theoretic features:  $\mathit{\Delta WML}$ is the difference between the WML values at $w_{n-1}$ and $w_{n}$; $\mathit{\Delta H}$ is the difference in entropy between  $w_{n-1}$ and $w_{n}$; $\mathit{InformationGain}$ is the difference between expected entropy at $w_{n-1}$ and observed $s$ at $w_n$, a measure that factors out the effect of naturally high entropy contexts; $\mathit{BestEntropyReduce}$ is the best reduction in entropy possible by an early rough hypothesis of reparandum onsets within 3 words; and $\mathit{BestWMLBoost}$ similarly speculates on the best improvement of $\mathit{WML}$ possible by positing $rm_{\mathit{start}}$ positions up to 3 words back. We also include simple alignment features: binary features which indicate if the word $w_{i-x}$ is identical to the current word $w_i$ for $x \in \{1,2,3\}$. With 6 alignment features, 16 N-gram features and a single logical feature $edit$ which indicates the presence of an edit word at position $w_{i-1}$, $rp_{\mathit{start}}$ detection uses 23 features-- see Table \ref{table:featuresRPSTART}.

We hypothesised repair onsets $rp^{\mathit{start}}$ would have significantly lower $p^{\mathit{lex}}$ (lower lexical-syntactic probability) and $\mathit{WML}^{\mathit{lex}}$ (lower syntactic probability) than other fluent trigrams. This was the case in the Switchboard heldout data for both measures, with the biggest difference obtained for $\mathit{WML}^{\mathit{lex}}$ (non-repair-onsets: -0.736 (sd=0.359); repair onsets: -1.457 (sd=0.359)). In the POS model, entropy of continuation $H^{\mathit{pos}}$ was the strongest feature (non-repair-onsets: 3.141 (sd=0.769); repair onsets: 3.444 (sd=0.899)). The trigram $\mathit{WML}^{\mathit{lex}}$ measure for the repaired utterance ``I haven't had any [ good + really very good ] experience with child care" can be seen in Figure~\ref{fig:WMLplot}. The steep drop at the repair onset shows the usefulness of  $\mathit{WML}$ features for fluency measures.

To compare n-gram measures against other local features, we ranked the features by Information Gain using 10-fold cross validation over the Switchboard heldout data-- see Table \ref{table:featuresRPSTART}. The language model features are far more discriminative than the alignment features, showing the potential of a general information-theoretic approach.

\subsubsection{Reparandum start detection}
In detecting $rm_{\mathit{start}}$ positions given a hypothesised $rp_{\mathit{start}}$ (stage T3 in Figure~\ref{fig:STIR}), we use the noisy channel intuition that removing the reparandum (from $rm_{\mathit{start}}$ to $rp_{\mathit{start}}$) increases fluency of the utterance, expressed here as $\mathit{WMLboost}$ as described above. When using gold standard input we found this was the case on the heldout data, with a mean $\mathit{WMLboost}$ of 0.223 (sd=0.267) for reparandum onsets and -0.058	(sd=0.224) for other words in the 6-word history- the negative boost for non-reparandum words captures the intuition that backtracking from those points would make the utterance less grammatical, and conversely the boost afforded by the correct $rm_{\mathit{start}}$ detection helps solve the continuation problem for the listener (and our detector). 

Parallelism in the onsets of $rp_{start}$ and $rm_{start}$ can also help solve the continuation problem, and in fact the KL divergence between $\theta^{pos}(w\given rm_{start},rm_{start-1})$ and $\theta^{pos}(w\given rp_{start},rp_{start-1})$  is the second most useful feature with average merit 0.429 (+- 0.010) in cross-validation. The highest ranked feature is $\mathit{\Delta WML}$ (0.437 (+- 0.003)) which here encodes the drop in the  $\mathit{WMLboost}$ from one backtracked position to the next. In ranking the 32 features we use, again information-theoretic ones are higher ranked than the logical features.

\begin{table}[h]
\begin{small}
\begin{tabular}{|c|c|c|}
\hline
average merit  & average rank &   attribute\\\hline
  0.139 (+- 0.002) &      1   (+- 0.00)  & $H^{\mathit{pos}}$  \\
 0.131 (+- 0.001)  &    2   (+- 0.00)   & $\mathit{WML}^{\mathit{pos}}$\\
 0.126 (+- 0.001)  &    3.4 (+- 0.66)   & $\mathit{WML}^{\mathit{lex}}$\\
 0.125 (+- 0.003)   &   4   (+- 1.10)    & $s^{\mathit{pos}}$\\
 0.122 (+- 0.001)  &    5.9 (+- 0.94)   &  $w_{i-1}=w_{i}$\\
 0.122 (+- 0.001)  &    5.9 (+- 0.70)    & BestWMLBoost$^{\mathit{lex}}$\\
 0.122 (+- 0.002)  &    5.9 (+- 1.22)   & InformationGain$^{\mathit{pos}}$\\
 0.119 (+- 0.001)  &    7.9 (+- 0.30)    & BestWMLBoost$^{\mathit{pos}}$\\
 0.098 (+- 0.002)  &    9   (+- 0.00)      & $H^{\mathit{lex}}$\\
 0.08  (+- 0.001) &    10.4 (+- 0.49)   & $\Delta \mathit{WML}^{\mathit{pos}}$\\
 0.08  (+- 0.003)  &   10.6 (+- 0.49)   & $\Delta H^{\mathit{pos}}$\\
 0.072 (+- 0.001)  &   12   (+- 0.00)     & $\mathit{POS}_{i-1}=\mathit{POS}_{i}$\\
 0.066 (+- 0.003) &    13.1 (+- 0.30)    &  $s^{\mathit{lex}}$\\
 0.059 (+- 0.000)  &      14.2 (+- 0.40)   &  $\Delta  \mathit{WML}^{\mathit{lex}}$\\
 0.058 (+- 0.005)  &   14.7 (+- 0.64)   & BestEntropyReduce$^{\mathit{pos}}$\\
 0.049 (+- 0.001) &    16.3 (+- 0.46)  & InformationGain$^{\mathit{lex}}$\\
 0.047 (+- 0.004) &    16.7 (+- 0.46)   &  BestEntropyReduce$^{\mathit{lex}}$\\
 0.035 (+- 0.004)  &   18   (+- 0.00)     &  $\Delta H^{\mathit{lex}}$\\
 0.024 (+- 0.000)  &       19   (+- 0.00)   &  $w_{i-2}=w_{i}$\\
 0.013 (+- 0.000)  &       20   (+- 0.00)    & $\mathit{POS}_{i-2}=\mathit{POS}_{i}$\\
 0.01  (+- 0.000)   &      21   (+- 0.00)    & $w_{i-3}=w_{i}$\\
 0.009 (+- 0.000)  &       22   (+- 0.00)   & edit\\
 0.006 (+- 0.000)  &       23   (+- 0.00)    & $\mathit{POS}_{i-3}=\mathit{POS}_{i}$\\
\hline
\end{tabular}
\end{small}
\caption{Feature ranker (Information Gain) for $rp_{\mathit{start}}$ detection- 10-fold x-validation on Switchboard heldout data.}\label{table:featuresRPSTART}
\end{table}

\subsubsection{Repair end detection and structure classification}

For $rp_{end}$ detection, using the notion of parallelism, we hypothesise an effect of divergence between $\theta^{\mathit{lex}}$ at the reparandum-final word $rm_{\mathit{end}}$ and the repair-final word $rp_{\mathit{end}}$: for repetition repairs,  KL divergence will trivially be 0; for substitutions, it will be higher; for deletes, even higher. Upon inspection of our feature ranking this KL measure ranked 5th out of 23 features (merit= 0.258 (+- 0.002)). 

We introduce another feature encoding parallelism $\mathit{ReparandumRepairDifference}$: the difference in probability between an utterance cleaned of the reparandum and the utterance with its repair phase substituting its reparandum. In both the POS (merit=0.366 (+- 0.003)) and word  (merit=0.352 (+- 0.002)) LMs, this was the most discriminative feature.

\subsection{Classifier pipeline}

\begin{figure*}
\begin{center}
 
   \includegraphics[width=\columnwidth]{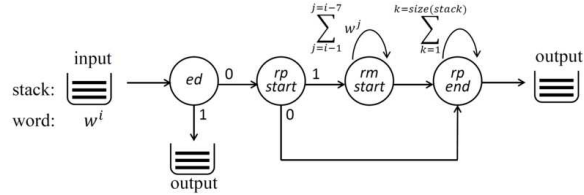} 
   
   \caption{Classifier pipeline}\label{fig:classifierPipeline}
   \end{center}
\end{figure*}

STIR effects a pipeline of classifiers as in Figure~\ref{fig:classifierPipeline}, where the $ed$ classifier only permits non $ed$ words to be passed on to $rp_{\mathit{start}}$ classification and for $rp_{\mathit{end}}$ classification of the active repair hypotheses, maintained in a stack. The $rp_{\mathit{start}}$ classifier passes positive repair hypotheses to the $rm_{\mathit{start}}$ classifier, which backwards searches up to 7 words back in the utterance. If a $rm_{\mathit{start}}$ is classified, the output is passed on for $rp_{\mathit{end}}$ classification at the end of the pipeline, and if not rejected this is pushed onto the repair stack. Repair hypotheses are are popped off when the string is 7 words beyond its $rp_{\mathit{start}}$ position. Putting limits on the stack's storage space is a way of controlling for processing overhead and complexity. Embedded repairs whose $rm_{\mathit{start}}$ coincide with another's $rp_{\mathit{start}}$ are easily dealt with as they are added to the stack as separate hypotheses.\footnote{We constrain the problem not to include embedded deletes which may share their $rp_{\mathit{start}}$ word with another repair -- these are in practice very rare.}

\paragraph{Classifiers}
Classifiers are implemented using Random Forests \cite{Breiman01} and we use different error functions for each stage using MetaCost \cite{Domingos99}. The flexibility afforded by implementing adjustable error functions in a pipelined incremental processor allows control of the trade-off of immediate accuracy against run-time and stability of the sequence classification.

\paragraph{Processing complexity} This pipeline avoids an exhaustive search all repair hypotheses. If we limit the search to within the $\langle rm_{start},rp_{start}\rangle$ possibilities, this number of repairs grows approximately in the triangular number series-- i.e. $\frac{n(n+1)}{2}$, a nested loop over previous words as $n$ gets incremented -- which in terms of a complexity class is a quadratic $O(n^{2})$. If we allow more than one $\langle rm_{start},rp_{start}\rangle$ hypothesis per word, the complexity goes up to $O(n^{3})$, however in the tests that we describe below, we are able to achieve good detection results without permitting this extra search space. Under our assumption that reparandum onset detection is only triggered after repair onset detection, and repair extent detection is dependent on positive reparandum onset detection, a pipeline with accurate components will allow us to limit processing to a small subset of this search space.


\section{Experimental set-up}
\label{sec:experiment}

We train STIR on the Switchboard data described above, and test it on the
standard Switchboard test data (PTB III files 4[0-1]*).  In order to
avoid over-fitting of classifiers to the basic language models, we use
a cross-fold training approach: we divide the corpus into 10 folds and
use language models trained on 9 folds to obtain feature values for
the 10th fold, repeating for all 10. Classifiers are then trained as
standard on the resulting feature-annotated corpus. This resulted in
better feature utility for n-grams and better F-score results for
detection in all components in the order of 5-6\%.\footnote{\newcite{Zwarts.Johnson11LMs} take a similar
  approach on Switchboard data  to train a re-ranker of repair  analyses.}

\paragraph{Training the classifiers} Each Random Forest classifier was limited to 20 trees of maximum depth 4 nodes, putting a ceiling on decoding time. In making the classifiers cost-sensitive, MetaCost resamples the data in accordance with the cost functions: we found using 10 iterations over a re-sample of 25\% of the training data gave the most effective trade-off between training time and accuracy.\footnote{As \cite{Domingos99} demonstrated, there are only relatively small accuracy gains when using more than this, with training time increasing in the order of the re-sample size.} We use 8 different cost functions in $rp_{\mathit{start}}$ with differing costs for false negatives and positives of the form below, where $R$ is a repair element word and $F$ is a fluent onset:

\begin{footnotesize}
\vspace{-0.2cm}
\begin{equation*}
\bordermatrix{~ & R^{hyp} & F^{hyp} \cr
                  R^{gold} & 0 & 2 \cr
                  F^{gold} & 1 & 0 \cr}
\end{equation*}
\vspace{-0.2cm}
\end{footnotesize}

We adopt a similar technique in $rm_{\mathit{start}}$ using 5 different cost functions and in $rp_{\mathit{end}}$ using 8 different settings, which when combined gives a total of 320 different cost function configurations. We hypothesise that higher recall permitted in the pipeline's first components would result in better overall accuracy as these hypotheses become refined, though at the cost of the stability of the hypotheses of the sequence and extra downstream processing in pruning false positives. 

We also experiment with the number of repair hypotheses permitted per word, using limits of 1-best and 2-best hypotheses. We expect that allowing 2 hypotheses to be explored per $rp_{\mathit{start}}$ should allow greater final accuracy, but with the trade-off of greater decoding and training complexity, and possible incremental instability.

As we wish to explore the incrementality versus final accuracy trade-off that STIR can achieve we now describe the evaluation metrics we employ.

\subsection{Incremental evaluation metrics}
\label{sec:incrementalEval}
Following \cite{Baumann.etal11evaluation} we divide our evaluation metrics into \emph{similarity metrics} (measures of equality with or similarity to a gold standard), \emph{timing metrics} (measures of the timing of relevant phenomena
detected from the gold standard) and \emph{diachronic metrics} (evolution of incremental hypotheses over time).

\paragraph{Similarity metrics} For direct comparison to previous approaches we use the standard measure of overall accuracy, the F-score over reparandum words, which we abbreviate \textbf{F$_{\mathbf{rm}}$} (see  \ref{eq:JohnsonCharniakEval}):

\vspace{-0.3cm}
\begin{small}
\begin{align}\label{eq:JohnsonCharniakEval}
\begin{split}
\text{precision} & =\frac{rm^{correct}}{rm^{hyp}}\\
\text{recall} & =\frac{rm^{correct}}{rm^{gold}}\\
\text{F$_{rm}$} & = 2 \times \frac{\text{precision} \times \text{recall}}{\text{precision} + \text{recall}}
\end{split}
\end{align}
\end{small}
\vspace{-0.3cm}

We are also interested in repair structural classification, we also measure F-score over \emph{all} repair components ($rm$ words, $ed$ words as interregna and $rp$ words), a metric we abbreviate \textbf{F$_{\mathbf{s}}$}. This is not measured in standard repair detection on Switchboard. To investigate incremental accuracy we evaluate the \emph{delayed accuracy} (\textbf{DA}) introduced by \cite{Zwarts.etal10}, as described in section \ref{sec:previous} against the utterance-final gold standard disfluency annotations, and use the mean of the 6 word F-scores.

\begin{figure}
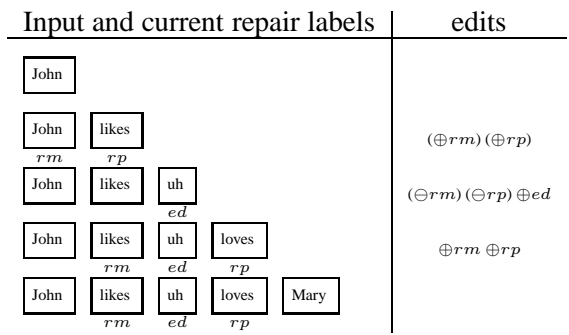

\begin{tiny}
\begin{tabular}{l|c}
\begin{normalsize}Input and current repair labels \end{normalsize}&      \begin{normalsize}edits\end{normalsize}\\\hline
\begin{tabular}{@{}c@{\hspace{0.2cm}}c@{}}& \\ \fbox{\strut John } & \\ & \end{tabular}         
\\
\begin{tabular}{@{}c@{\hspace{0.2cm}}c@{}}\fbox{\strut John }&\fbox{\strut likes }\\ $\strut rm$ & $\strut rp$ \end{tabular}               &                              
($\oplus rm$) ($\oplus rp$)
\\
\begin{tabular}{@{}c@{\hspace{0.2cm}}c@{\hspace{0.2cm}}c@{}}\fbox{\strut John }&\fbox{\strut likes } & \fbox{\strut uh }\\
 &   &  $\strut  ed$ \end{tabular}    &                                       
 ($\ominus rm$)  ($\ominus rp$) $\oplus ed$
\\
\begin{tabular}{@{}c@{\hspace{0.2cm}}c@{\hspace{0.2cm}}c@{\hspace{0.2cm}}c@{}}\fbox{\strut John }&\fbox{\strut likes } & \fbox{\strut uh } &\fbox{\strut loves }\\
 & $rm$  &  $\strut  ed$ & $rp$ \end{tabular}    &         
 $\oplus rm$ $\oplus rp$
\\
\begin{tabular}{@{}c@{\hspace{0.2cm}}c@{\hspace{0.2cm}}c@{\hspace{0.2cm}}c@{\hspace{0.2cm}}c@{}}\fbox{\strut John }&\fbox{\strut likes } & \fbox{\strut uh } &\fbox{\strut loves } & \fbox{\strut Mary }\\
 & $rm$  &  $\strut  ed$ & $rp$ &  \end{tabular}    &         
\\
\end{tabular}
\end{tiny}
\caption{Edit Overhead- 4 unnecessary  edits}\label{fig:editOverhead}
\end{figure}

\paragraph{Timing and resource metrics} Again for comparative purposes we use Zwarts et al's \emph{time-to-detection} metrics, that is the two average distances (in numbers of words) consumed before first detection of gold standard repairs, one from $rm_{\mathit{start}}$, \textbf{TD$_{\mathbf{rm}}$} and one from $rp_{\mathit{start}}$, \textbf{TD$_{\mathbf{rp}}$}. In our 1-best detection system, before evaluation we know a priori TD$_{rp}$ will be 1 token, and TD$_{rm}$ will be 1 more than the average length of $rm_{start}-rp_{start}$ repair spans correctly detected. However when we introduce a beam where multiple $rm_{start}$s are possible per $rp_{start}$ with the most likely hypothesis committed as the current output, the latency may begin to increase: the initially most probable hypothesis may not be the correct one. In addition to output timing metrics, we account for intrinsic processing complexity with the metric \emph{processing overhead} (\textbf{PO}), which is the number of classifications made by all components per word of input.

\paragraph{Diachronic metrics} To measure stability of repair hypotheses over time we use \cite{Baumann.etal11evaluation}'s \emph{edit overhead} (\textbf{EO}) metric. EO measures the proportion of edits (add, revoke, substitute) applied to a processor's output structure that are unnecessary. STIR's output is the repair label sequence shown in Figure \ref{fig:STIR}, however rather than evaluating its EO against the current gold standard labels, we use a new mark-up we term the \emph{incremental repair gold standard}: this does not penalise lack of detection of a reparandum word $rm$ as a bad edit until the corresponding $rp_{\mathit{start}}$ of that $rm$ has been consumed. While F$_{rm}$, F$_s$ and DA evaluate against what \newcite{Baumann.etal11evaluation} call the \emph{current gold standard}, the incremental gold standard reflects the repair processing approach we set out in \ref{sec:approach}. An example of a repaired utterance with an EO of 44\% ($\frac{4}{9}$) can be seen in Figure~\ref{fig:editOverhead}: of the 9 edits (7 repair annotations and 2 correct fluent words), 4 are unnecessary (bracketed). Note the final $\oplus rm$ is not counted as a bad edit for the reasons just given.

\begin{figure*}[!t]
\begin{footnotesize}
\begin{tabular}{cccc}
$
\bordermatrix{~ & rp_{start}^{hyp} & F^{hyp} \cr
                  rp_{start}^{gold} & 0 & 64 \cr
                  F^{gold} & 1 & 0 \cr}
$&  
$
\bordermatrix{~ & rm_{start}^{hyp} & F^{hyp} \cr
                  rm_{start}^{gold} & 0 & 8 \cr
                  F^{gold} & 1 & 0 \cr}
$& 
$
\bordermatrix{~ & rp_{end}^{hyp} & F^{hyp} \cr
                  rp_{end}^{gold} & 0 & 2 \cr
                  F^{gold} & 1 & 0 \cr}
$& 
Stack depth = 2\\$
\bordermatrix{~ & rp_{start}^{hyp} & F^{hyp} \cr
                  rp_{start}^{gold} & 0 & 2 \cr
                  F^{gold} & 1 & 0 \cr}
$&  
$
\bordermatrix{~ & rm_{start}^{hyp} & F^{hyp} \cr
                  rm_{start}^{gold} & 0 & 16 \cr
                  F^{gold} & 1 & 0 \cr}
$& 
$
\bordermatrix{~ & rp_{end}^{hyp} & F^{hyp} \cr
                  rp_{end}^{gold} & 0 & 8 \cr
                  F^{gold} & 1 & 0 \cr}
$& 
Stack depth = 1\\
\end{tabular}
\end{footnotesize}
\caption{The cost function settings for the MetaCost classifiers for each component, for the best F$_{rm}$ setting (top row) and best total score (TS) setting (bottom row)}\label{fig:costFunctions}
\end{figure*}

\begin{table*}\centering
\begin{tabular}{lccccc}
									&	F$_{rm}$& F$_s$ & DA & EO & PO \\
									
									\hline
Best Final $rm$ F-score  (F$_{rm}$) & \textbf{0.779}	& 0.735	& 0.698 & 	3.946 & 	1.733 \\
Best Final repair structure F-score  (F$_s$) & 0.772	&  \textbf{0.736} &	0.707 &	4.477 & 1.659 \\
Best Delayed Accuracy of $rm$	(DA) & 0.767 &	0.721	& \textbf{0.718} &	1.483 & 1.689 \\
Best (lowest) Edit Overhead (EO) & 0.718 & 0.674	& 0.675	 & \textbf{0.864} & 1.230  \\
Best (lowest) Processing Overhead (PO) & 0.716 &	0.671	& 0.673 &	0.875 & \textbf{1.229} \\
Best Total Score (mean \% of best scores) (TS) & \emph{0.754} &	\emph{0.708} & \emph{0.711}&\emph{0.931} & \emph{1.255}  \\
\end{tabular}
\caption{Comparison of the best performing system settings using different measures}\label{table:overallresults}
\end{table*}

\begin{table*}[!t] \centering
\begin{tabular}{lccccccc}
 & F$_{rm}$ & F$_s$ & DA & EO & PO & TD$_{rp}$ & TD$_{rm}$ \\\hline
1-best $rm_{start}$	& 0.745 &	0.707 &	0.699 &3.780  & 1.650 & 1.0 & 2.6 \\
2-best $rm_{start}$ 	 & 0.758 &	0.721& 0.701 & 4.319	&  1.665& 1.1 & 2.7 \\
\end{tabular}
\caption{Comparison of performance of systems with different stack capacities}\label{table:stack}
 \end{table*}

\section{Results and Discussion}
\label{sec:results}

We evaluate on the Switchboard test data; Table~\ref{table:overallresults} shows results of the best performing settings for each of the 
metrics described above, together with the setting achieving the highest total score (\textbf{TS})-- the average \% achieved of the best performing system's result in each metric.\footnote{We do not include time-to-detection scores in TS as it did not vary enough between settings to be significant, however there was a difference in this measure between the 1-best stack condition and the 2-best stack condition -- see below.} The settings found to achieve the highest F$_{\mathbf{rm}}$ (the metric standardly used in disfluency detection), and that found to achieve the highest TS for each stage in the pipeline are shown in Figure~\ref{fig:costFunctions}.

\begin{figure}
\begin{center}
 \includegraphics[width=\columnwidth]{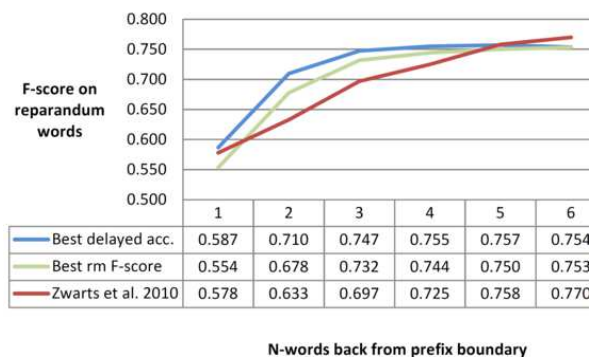} 
   
   \caption[Delayed Accuracy]{Delayed Accuracy Curves}\label{fig:delayedAccuracy}
   \end{center}
\end{figure}

Our experiments showed that different system settings perform better in different metrics, and no individual setting achieved the best result in all of them. Our best utterance-final F$_{rm}$ reaches 0.779, marginally though not significantly exceeding \cite{Zwarts.etal10}'s measure and STIR achieves 0.736 on the previously unevaluated F$_{s}$. The setting with the best DA improves on \cite{Zwarts.etal10}'s result significantly in terms of mean values (0.718 vs. 0.694), and also in terms of the steepness of the curves (Figure \ref{fig:delayedAccuracy}). The fastest average time to detection is 1 word for TD$_{rp}$ and 2.6 words for TD$_{rm}$ (Table~\ref{table:stack}), improving dramatically on the noisy channel model's 4.6 and 7.5 words.

\paragraph{Incrementality versus accuracy trade-off} We aimed to investigate how well a system could do in terms of achieving both good final accuracy and incremental performance, and while the best F$_{rm}$ setting had a large PO and relatively slow DA increase, we find STIR can find a good trade-off setting:  the highest TS scoring setting achieves an F$_{rm}$ of 0.754 whilst also exhibiting a very good DA (0.711) -- over 98\% of the best recorded score --  and low PO and EO rates -- over 96\% of the best recorded scores. See the bottom row of Table~\ref{table:overallresults}. As can be seen in Figure~\ref{fig:costFunctions},  the cost functions for these winning settings are different in nature. The best non-incremental F$_{rm}$ measure setting requires high recall for the rest of the pipeline to work on, using the highest cost, 64, for false negative $rp_{\mathit{start}}$ words and the highest stack depth of 2 (similar to a wider beam); but the best overall TS scoring system uses a less permissive setting to increase incremental performance.

We make a preliminary investigation into the effect of increasing the stack capacity by comparing stacks with 1-best $rm_{start}$ hypotheses per $rp_{start}$ and 2-best stacks. The average differences between the two conditions is shown in Table \ref{table:stack}. Moving to the 2-stack condition results in gain in overall accuracy in F$_{rm}$ and F$_s$, but at the cost of EO and also time-to-detection scores TD$_{rm}$ and TD$_{rp}$. The extent to which the stack can be increased without increasing jitter, latency and complexity will be investigated in future work.

\section{Conclusion}
\label{sec:conclusion}
We have presented STIR, an incremental repair detector that can be used to experiment with incremental performance and accuracy trade-offs. In future work we plan to include probabilistic and distributional features from a top-down incremental parser e.g. \newcite{Roark.etal09}, and use STIR's distributional features to classify repair type.

\section*{Acknowledgements}

We thank the three anonymous EMNLP reviewers for their helpful comments.
Hough is supported by the DUEL project, financially supported by the Agence Nationale de la Research (grant number ANR-13-FRAL-0001) and the Deutsche Forschungsgemainschaft. Much of the work was carried out with support from an EPSRC DTA scholarship at Queen Mary University of London.
Purver is partly supported by ConCreTe: the project ConCreTe
acknowledges the financial support of the Future and Emerging
Technologies (FET) programme within the Seventh Framework Programme
for Research of the European Commission, under FET grant number
611733.


\bibliographystyle{acl}
\bibliography{../../all}

\begin{thebibliography}{}

\bibitem[\protect\citename{Baumann \bgroup et al.\egroup
  }2011]{Baumann.etal11evaluation}
T.~Baumann, O.~Bu{\ss}, and D.~Schlangen.
\newblock 2011.
\newblock Evaluation and optimisation of incremental processors.
\newblock {\em Dialogue \& Discourse}, 2(1):113--141.

\bibitem[\protect\citename{Breiman}2001]{Breiman01}
Leo Breiman.
\newblock 2001.
\newblock Random forests.
\newblock {\em Machine learning}, 45(1):5--32.

\bibitem[\protect\citename{Brennan and Schober}2001]{Brennan.Schober01}
S.E. Brennan and M.F. Schober.
\newblock 2001.
\newblock How listeners compensate for disfluencies in spontaneous speech.
\newblock {\em Journal of Memory and Language}, 44(2):274--296.

\bibitem[\protect\citename{Brill and Moore}2000]{Brill.Moore00}
Eric Brill and Robert~C Moore.
\newblock 2000.
\newblock An improved error model for noisy channel spelling correction.
\newblock In {\em Proceedings of the 38th Annual Meeting on Association for
  Computational Linguistics}, pages 286--293. Association for Computational
  Linguistics.

\bibitem[\protect\citename{Clark \bgroup et al.\egroup }2013]{Clark.etal13CMCL}
Alexander Clark, Gianluca Giorgolo, and Shalom Lappin.
\newblock 2013.
\newblock Statistical representation of grammaticality judgements: the limits
  of n-gram models.
\newblock In {\em Proceedings of the Fourth Annual Workshop on Cognitive
  Modeling and Computational Linguistics (CMCL)}, pages 28--36, Sofia,
  Bulgaria, August. Association for Computational Linguistics.

\bibitem[\protect\citename{Domingos}1999]{Domingos99}
Pedro Domingos.
\newblock 1999.
\newblock Metacost: A general method for making classifiers cost-sensitive.
\newblock In {\em Proceedings of the fifth ACM SIGKDD international conference
  on Knowledge discovery and data mining}, pages 155--164. ACM.

\bibitem[\protect\citename{Georgila}2009]{Georgila09}
Kallirroi Georgila.
\newblock 2009.
\newblock Using integer linear programming for detecting speech disfluencies.
\newblock In {\em Proceedings of Human Language Technologies: The 2009 Annual
  Conference of the North American Chapter of the Association for Computational
  Linguistics, Companion Volume: Short Papers}, pages 109--112. Association for
  Computational Linguistics.

\bibitem[\protect\citename{Germesin \bgroup et al.\egroup
  }2008]{Germesin.etal08}
Sebastian Germesin, Tilman Becker, and Peter Poller.
\newblock 2008.
\newblock Hybrid multi-step disfluency detection.
\newblock In {\em Machine Learning for Multimodal Interaction}, pages 185--195.
  Springer.

\bibitem[\protect\citename{Ginzburg}2012]{Ginzburg12}
Jonathan Ginzburg.
\newblock 2012.
\newblock {\em The Interactive Stance: Meaning for Conversation}.
\newblock Oxford University Press.

\bibitem[\protect\citename{Healey \bgroup et al.\egroup }2011]{Healey.etal11}
P.~G.~T. Healey, Arash Eshghi, Christine Howes, and Matthew Purver.
\newblock 2011.
\newblock Making a contribution: Processing clarification requests in dialogue.
\newblock In {\em Proceedings of the 21st Annual Meeting of the Society for
  Text and Discourse}, Poitiers, July.

\bibitem[\protect\citename{Heeman and Allen}1999]{Heeman.Allen99}
Peter Heeman and James Allen.
\newblock 1999.
\newblock Speech repairs, intonational phrases, and discourse markers: modeling
  speakers' utterances in spoken dialogue.
\newblock {\em Computational Linguistics}, 25(4):527--571.

\bibitem[\protect\citename{Honnibal and Johnson}2014]{Honnibal.Johnson14}
Matthew Honnibal and Mark Johnson.
\newblock 2014.
\newblock Joint incremental disfluency detection and dependency parsing.
\newblock {\em Transactions of the Association of Computational Linugistics
  (TACL)}, 2:131--142.

\bibitem[\protect\citename{Hough and Purver}2013]{Hough.Purver13SemDial}
Julian Hough and Matthew Purver.
\newblock 2013.
\newblock Modelling expectation in the self-repair processing of annotat-, um,
  listeners.
\newblock In {\em Proceedings of the 17th {SemDial} Workshop on the Semantics
  and Pragmatics of Dialogue ({DialDam})}, pages 92--101, Amsterdam, December.

\bibitem[\protect\citename{Jaeger and Tily}2011]{Jaeger.Tily11}
T~Florian Jaeger and Harry Tily.
\newblock 2011.
\newblock On language ‘utility’: Processing complexity and communicative
  efficiency.
\newblock {\em Wiley Interdisciplinary Reviews: Cognitive Science},
  2(3):323--335.

\bibitem[\protect\citename{Johnson and Charniak}2004]{Johnson.Charniak04}
Mark Johnson and Eugene Charniak.
\newblock 2004.
\newblock A {TAG}-based noisy channel model of speech repairs.
\newblock In {\em Proceedings of the 42nd Annual Meeting on Association for
  Computational Linguistics}, pages 33--39, Barcelona. Association for
  Computational Linguistics.

\bibitem[\protect\citename{Keller}2004]{Keller04}
Frank Keller.
\newblock 2004.
\newblock The entropy rate principle as a predictor of processing effort: An
  evaluation against eye-tracking data.
\newblock In {\em EMNLP}, pages 317--324.

\bibitem[\protect\citename{Kneser and Ney}1995]{Kneser.Ney95}
Reinhard Kneser and Hermann Ney.
\newblock 1995.
\newblock Improved backing-off for m-gram language modeling.
\newblock In {\em Acoustics, Speech, and Signal Processing, 1995. ICASSP-95.,
  1995 International Conference on}, volume~1, pages 181--184. IEEE.

\bibitem[\protect\citename{Lease \bgroup et al.\egroup }2006]{Lease.etal06}
Matthew Lease, Mark Johnson, and Eugene Charniak.
\newblock 2006.
\newblock Recognizing disfluencies in conversational speech.
\newblock {\em Audio, Speech, and Language Processing, IEEE Transactions on},
  14(5):1566--1573.

\bibitem[\protect\citename{Levelt}1983]{Levelt83}
W.J.M. Levelt.
\newblock 1983.
\newblock Monitoring and self-repair in speech.
\newblock {\em Cognition}, 14(1):41--104.

\bibitem[\protect\citename{Meteer \bgroup et al.\egroup }1995]{Meeter.etal95}
M.~Meteer, A.~Taylor, R.~MacIntyre, and R.~Iyer.
\newblock 1995.
\newblock Disfluency annotation stylebook for the switchboard corpus. ms.
\newblock Technical report, Department of Computer and Information Science,
  University of Pennsylvania.

\bibitem[\protect\citename{Miller and Schuler}2008]{Miller.Schuler08}
Tim Miller and William Schuler.
\newblock 2008.
\newblock A syntactic time-series model for parsing fluent and disfluent
  speech.
\newblock In {\em Proceedings of the 22nd International Conference on
  Computational Linguistics-Volume 1}, pages 569--576. Association for
  Computational Linguistics.

\bibitem[\protect\citename{Milward}1991]{Milward91}
David Milward.
\newblock 1991.
\newblock {\em Axiomatic Grammar, Non-Constituent Coordination and Incremental
  Interpretation}.
\newblock {Ph.D.} thesis, University of Cambridge.

\bibitem[\protect\citename{Qian and Liu}2013]{Qian.Liu13}
Xian Qian and Yang Liu.
\newblock 2013.
\newblock Disfluency detection using multi-step stacked learning.
\newblock In {\em Proceedings of NAACL-HLT}, pages 820--825.

\bibitem[\protect\citename{Rasooli and Tetreault}2014]{Rasooli.Tetreault14}
Mohammad~Sadegh Rasooli and Joel Tetreault.
\newblock 2014.
\newblock Non-monotonic parsing of fluent umm {I} mean disfluent sentences.
\newblock {\em EACL 2014}, pages 48--53.

\bibitem[\protect\citename{Rieser and Schlangen}2011]{Rieser.Schlangen11}
Hannes Rieser and David Schlangen.
\newblock 2011.
\newblock Introduction to the special issue on incremental processing in
  dialogue.
\newblock {\em Dialogue \& Discourse}, 2(1):1--10.

\bibitem[\protect\citename{Roark \bgroup et al.\egroup }2009]{Roark.etal09}
Brian Roark, Asaf Bachrach, Carlos Cardenas, and Christophe Pallier.
\newblock 2009.
\newblock Deriving lexical and syntactic expectation-based measures for
  psycholinguistic modeling via incremental top-down parsing.
\newblock In {\em Proceedings of the 2009 Conference on Empirical Methods in
  Natural Language Processing: Volume 1-Volume 1}, pages 324--333. Association
  for Computational Linguistics.

\bibitem[\protect\citename{Shannon}1948]{Shannon48}
Claude~E. Shannon.
\newblock 1948.
\newblock A mathematical theory of communication. technical journal.
\newblock {\em AT \& T Bell Labs}.

\bibitem[\protect\citename{Shriberg and Stolcke}1998]{Shriberg.Stolcke98}
Elizabeth Shriberg and Andreas Stolcke.
\newblock 1998.
\newblock How far do speakers back up in repairs? {A} quantitative model.
\newblock In {\em Proceedings of the International Conference on Spoken
  Language Processing}, pages 2183--2186.

\bibitem[\protect\citename{Shriberg}1994]{Shriberg94}
Elizabeth Shriberg.
\newblock 1994.
\newblock {\em Preliminaries to a Theory of Speech Disfluencies}.
\newblock {Ph.D.} thesis, University of California, Berkeley.

\bibitem[\protect\citename{Zwarts and Johnson}2011]{Zwarts.Johnson11LMs}
Simon Zwarts and Mark Johnson.
\newblock 2011.
\newblock The impact of language models and loss functions on repair disfluency
  detection.
\newblock In {\em Proceedings of the 49th Annual Meeting of the Association for
  Computational Linguistics: Human Language Technologies - Volume 1}, HLT '11,
  pages 703--711, Stroudsburg, PA, USA. Association for Computational
  Linguistics.

\bibitem[\protect\citename{Zwarts \bgroup et al.\egroup }2010]{Zwarts.etal10}
Simon Zwarts, Mark Johnson, and Robert Dale.
\newblock 2010.
\newblock Detecting speech repairs incrementally using a noisy channel
  approach.
\newblock In {\em Proceedings of the 23rd International Conference on
  Computational Linguistics}, COLING '10, pages 1371--1378, Stroudsburg, PA,
  USA. Association for Computational Linguistics.

\end{thebibliography}

\end{document}